\documentclass{ieeeaccess}
\usepackage{cite}
\usepackage{amsmath,amssymb,amsfonts}
\usepackage{graphicx}
\usepackage{textcomp}
\usepackage{algorithm}
\usepackage{listings}
\usepackage{lipsum}
\usepackage{multirow}
\usepackage{lineno,hyperref}
\usepackage{algpseudocode}
\usepackage{booktabs}
\usepackage{mathtools}
\usepackage{upgreek}
\usepackage{soul,color}
\usepackage{hyperref}
\usepackage{eurosym}
\usepackage{siunitx}
\usepackage{textgreek}
\soulregister\cite7
\soulregister\ref7
\soulregister\sc7
\def\textsc#1{\textnormal{{\sc #1}}}
\usepackage{caption}
\usepackage{subcaption}
\DeclareSIUnit{\EUR}{\text{\euro}}
\DeclareSIUnit{\kgc}{kg\ CO\textsubscript{2}}
\DeclareSIUnit{\km}{km}
\DeclareSIUnit{\liter}{L}
\DeclareSIUnit{\kwh}{kWh}
\DeclareCaptionFont{ieeeblue}{\color{accessblue}}
\DeclareCaptionLabelFormat{myformat}{\raggedright\figcapfont{\textbf{#1}\textbf{#2}}}
\def\BibTeX{{\rm B\kern-.05em{\sc i\kern-.025em b}\kern-.08em
T\kern-.1667em\lower.7ex\hbox{E}\kern-.125emX}}

\newcommand{\expnumber}[2]{{#1}\text{e}\textsuperscript{#2}}
\newcommand{\sctxt}[1]{{\sc #1}}
\newcommand{\red}[1]{{\color{red} #1}}
\soulregister{\sctxt}{7}
\soulregister{\red}{7}
\soulregister{\expnumber}{7}
\soulregister{\SI}{7}
\soulregister{\percent}{7}

\begin{document}

\history{Received 16 November 2022, accepted 30 November 2022, date of publication 1 December 2022, date of current version 7 December 2022.}
\doi{10.1109/ACCESS.2022.3226324}

\title{Explainable automatic industrial carbon footprint estimation from bank transaction classification using natural language processing}

\author{\uppercase{Jaime González-González}\authorrefmark{1},\uppercase{Silvia García-Méndez}\authorrefmark{1},\uppercase{Francisco de Arriba-Pérez}\authorrefmark{1},\uppercase{ Francisco J. Gonz\'alez-Casta\~no}\authorrefmark{1},\uppercase{ and Óscar Barba-Seara}\authorrefmark{2}}
\address[1]{Information Technologies Group, atlanTTic, University of Vigo, Telecommunication Engineering School, Campus, 36310 Vigo, Spain}
\address[2]{CoinScrap Finance S.L., 36002 Pontevedra, Spain}
\tfootnote{This work was partially supported by Xunta de Galicia grants ED481B-2021-118 and ED481B-2022-093, Spain; and Spanish national funds through \textit{Centro para el Desarrollo Tecnológico Industrial} (Center for Industrial Technological Development) as part of the project EXP00146826/IDI-20220298. The authors are indebted to CoinScrap Finance S.L. for providing us with the experimental data-set.}

\markboth
{Jaime González-González \headeretal: Explainable automatic industrial carbon footprint estimation from bank transaction classification using NLP}
{Jaime González-González \headeretal: Explainable automatic industrial carbon footprint estimation from bank transaction classification using NLP}

\corresp{Corresponding author: Silvia García-Méndez (e-mail: sgarcia@gti.uvigo.es).}

\begin{abstract}
Concerns about the effect of greenhouse gases have motivated the development of certification protocols to quantify the industrial carbon footprint (\sctxt{cf}). These protocols are manual, work-intensive, and expensive. All of the above have led to a shift towards automatic data-driven approaches to estimate the \sctxt{cf}, including Machine Learning (\sctxt{ml}) solutions. Unfortunately, as in other sectors of interest, the decision-making processes involved in these solutions lack transparency from the end user's point of view, who must blindly trust their outcomes compared to intelligible traditional manual approaches. In this research, manual and automatic methodologies for \textsc{cf} estimation were reviewed, taking into account their transparency limitations. This analysis led to the proposal of a new explainable \sctxt{ml} solution for automatic \sctxt{cf} calculations through bank transaction classification. Consideration should be given to the fact that no previous research has considered the explainability of bank transaction classification for this purpose. For classification, different \textsc{ml} models have been employed based on their promising performance in similar problems in the literature, such as Support Vector Machine, Random Forest, and Recursive Neural Networks. The results obtained were in the 90 \% range for accuracy, precision, and recall evaluation metrics. From their decision paths, the proposed solution estimates the \sctxt{co\textsubscript{2}} emissions associated with bank transactions. The explainability methodology is based on an agnostic evaluation of the influence of the input terms extracted from the descriptions of transactions using locally interpretable models. The explainability terms were automatically validated using a similarity metric over the descriptions of the target categories. Conclusively, the explanation performance is satisfactory in terms of the proximity of the explanations to the associated activity sector descriptions, endorsing the trustworthiness of the process for a human operator and end users.

\end{abstract}


\titlepgskip=-15pt

\maketitle

\section{Introduction}
\label{introduction}

\subsection{Research gap and motivation}

Concerns about climatic change\cite{SDG2021,IPCC2018} related to the increasing emission of greenhouse gases (\sctxt{ghg}) led 187 countries to sign the Paris Agreement\footnote{Available at \url{https://unfccc.int/process-and-meetings/the-paris-agreement/the-paris-agreement}, November 2022.} in 2015. This accord expressed the need for policies and regulations on \sctxt{ghg} emissions such as carbon dioxide (\sctxt{co\textsubscript{2}}). The so-called carbon footprint (\sctxt{cf}) can be defined as the amount of \sctxt{ghg} released to the atmosphere throughout the life cycle of a product or human activity\cite{Wiedmann2008}. Over the years, there have been many proposals to estimate the \sctxt{cf} of different entities\cite{Pandey2011}, including individuals, families, industries, and geographical bodies such as cities\cite{Ionescu2022}. 

The motivations for the calculation of \sctxt{cf} are diverse, with compliance with environmental legislation and the certification of industrial sustainability (\sctxt{iso} 14064\footnote{Available at \url{https://www.iso.org/standard/66453.html}, November 2022.}) being two of the most relevant reasons. Another relevant inducement is self-checking to avoid environmental taxes\cite{Zhu2020} and attract funding from ecologically-minded investors\cite{Ionescu2021}. Moreover, individuals, especially young people, have pressing concerns regarding the effects of climate change\cite{Milfont2012, Luis2018}. Consequently, diverse tracking applications allow end users to estimate and reduce their \textsc{cf}\cite{Hoffmann2022}.

\sctxt{cf} estimation solutions can be divided into manual and automatic approaches:

\begin{description}

 \item \textbf{Manual solutions.} For individuals, manual calculator applications require estimates of consumption habits, travel, \textit{etc.}, as input data. These applications employ predefined formulae\cite{Mulrow2019}. For industrial certifications, there exist consulting companies, such as \sctxt{aecom}\footnote{Available at \url{https://aecom.com/services/environmental-services}, November 2022.} and \sctxt{kpmg}\footnote{Available at \url{https://home.kpmg/xx/en/home/insights/2020/12/environmental-social-governance-esg-and-sustainability.html}, November 2022.} whose environmental services include \sctxt{cf} estimation.
 
 \item \textbf{Automatic solutions.} Some examples are the \textit{DO}\footnote{Available at \url{https://www.diva-portal.org/smash/get/diva2:1604075/FULLTEXT01.pdf}, November 2022.}, \textit{Enfuce}\footnote{Available at \url{https://enfuce.com}, November 2022.} and \textit{Joro}\footnote{Available at \url{https://www.joro.app}, November 2022.} apps. Supervised approaches rely on the Classification of Individual Consumption by Purpose (\sctxt{coicop}\footnote{Available at \url{https://unstats.un.org/unsd/class/revisions/coicop_revision.asp}, November 2022.}) by the United Nations or other categories of consumption habits. Bank transactions are useful for individuals useful\cite{Andersson2020}. For industries, little Enterprise Resource Planning (\sctxt{erp}) includes \textsc{cf} estimation\cite{Zvezdov2016}. 
 
\end{description}

As far as we know, although automatic estimation of \sctxt{cf} from bank transaction descriptions has already been considered for end users, it is a novel problem in the industry. In fact, the explainability of industrial \textsc{cf} estimation based on the automatic classification of bank transactions has not yet been considered in previous research, as supported by the state-of-the-art discussion in Section \ref{related_work}.

\subsection{Contribution}
\label{subcontribution}

In this paper, we propose an explainable automatic solution for industrial \sctxt{cf} estimation based on a supervised bank transaction classification model. The training set was labeled as \sctxt{coicop} classes. 

Regrettably, classification tasks performed by Machine Learning (\sctxt{ml}) models are often opaque\cite{Chandra2020}, which may affect customer trust, especially in industrial contexts; hence, there is a growing interest in Explainable Artificial Intelligence (\sctxt{xai}). Explainability methodologies allow for the extraction of intrinsic knowledge about the models' decisions\footnote{Available at \url{https://www.darpa.mil/program/explainable-artificial-intelligence}, November 2022.}.

Departing from a categorization model combining \sctxt{ml} with Natural Language Processing (\sctxt{nlp}) techniques, the main contribution of this study lies in the proposal of the automatic explainability of \sctxt{cf} estimation decisions. As previously mentioned, no authors have considered this aspect despite its relevance, for example, to examine consultancy analytics. The methodology extracts a set of relevant words for the classifier, and this word set is then validated with a similarity metric by comparing it with descriptions of the corresponding activity sectors.

\subsection{Paper organization}

The remainder of this paper is organized as follows. Section \ref{related_work} reviews the state of the art in bank transaction classification applied to \sctxt{cf} calculation using \sctxt{ml} models and focuses on the explainability feature. Section \ref{architecture} describes the proposed architecture for explainable automatic industrial \textsc{cf} estimation. Section \ref{evaluation} presents the experimental data-set and implementations used, along with the results obtained in terms of classification and explainability. Finally, Section \ref{conclusions} summarizes the conclusions and proposes future research.

\section{Related Work}
\label{related_work}

Many previous studies have applied \sctxt{ml} in fields such as E-commerce\cite{Tan2020}, incident management in information systems\cite{Silva2018}, and medical record analysis\cite{Berge2019}. In finance\cite{Huang2020, Rakshit2022}, \textsc{ml} models have been considered for detecting financial opportunities in social networks\cite{DeArriba-Perez2020}, fraud\cite{Ashitani2021, Kolli2020}, market sentiment\cite{Mishev2020}, risk\cite{Bhatore2020}, accounting\cite{Bardelli2020}, and financial transaction classification\cite{Jorgensen2021}.

In particular, bank transaction classification is a type of short-text classification that was already covered in our previous work\cite{Garcia-Mendez2020}. The latter topic has been applied to problems among those in which intelligent budget management deserves attention\cite{Folkestad2017,Vollset2017,Allegue2021}. 

Nevertheless, no previous work on bank transaction classification had an \sctxt{xai} perspective (with the sole exception of Kotios \textit{el al.} (2022)\cite{Kotios2022}, although it did not involve any \sctxt{nlp} methodology) nor considered industrial \sctxt{cf} estimation.

The base classification methodologies involved are numerous and include simple Naive Bayes classifiers\cite{Gao2019}, supervised learning models such as Random Forests (\sctxt{rf})\cite{Amir2019, Taskin2018}, and Support Vector Machine (\sctxt{svm})\cite{Goudjil2018, Kim2019}, along with more complex approaches based on Deep Learning (\sctxt{dl}) and Neural Networks (\sctxt{nn})\cite{Wang2019, Almuzaini2020}. 

The first solutions for \sctxt{cf} estimation typically follow official protocols and practices\footnote{Available at \url{https://ghgprotocol.org/standards}, November 2022.} and rely on manual calculations\cite{Adewale2019,Mulrow2019}. These protocols are time consuming and expensive to apply at the industrial level. More recent solutions oriented to end users have performed automatic \sctxt{cf} estimation from bank transactions\cite{Andersson2020} and employed social networking\cite{Biorn2020} to foster user engagement\cite{Barendregt2020}. 

End users are mainly motivated by environmental awareness and may be less concerned about the decision transparency of solutions. Conversely, industrial users may obtain important advantages from the application of automatic methodologies based on enterprise data, but solution transparency must be provided. In this regard, the incorporation of \sctxt{ai} in Industry 4.0 has boosted the application of \sctxt{xai} strategies in recent years\cite{Emmert2020,Ahmed2020} to shed some light on the decisions of automatically supervised\cite{Burkart2021} and unsupervised\cite{Montavon2022, Heuillet2021} learning models. Furthermore, explainability allows 
the prediction of behavior of these algorithms\cite{Gunning2019}. 

The existence of different explainability approaches is motivated by the variety of learning algorithms:

\begin{itemize}

 \item \textbf{Model-agnostic explainability}. It considers \sctxt{ml} models as black boxes and applies reverse engineering to infer their behavior.

     \begin{itemize}
     
     \item \textbf{Model induction}. It consists of a counterfactual study of feature changes\cite{Wachter2017} or a correlation analysis of features and outputs\cite{Goldstein2015, Apley2020}.
     
     \item \textbf{Local explanation}. It exploits local linear interpretable models that match the results of those under analysis\cite{Ribeiro2016, Plumb2018}. These local explanations can be enhanced using additional contextual or semantic information\cite{Kiefer2022}.
     
     \end{itemize}
     
 \item \textbf{Model-dependent explainability}. It is based on the inherent structure of \sctxt{ml} models.
 
     \begin{itemize}
     
     \item \textbf{Interpretable models}. Certain learning techniques are easily understandable to humans, as in the case of Decision Tree (\sctxt{dt})\cite{Sagi2020, Cousins2019, Ozge2021}, \sctxt{rf}\cite{Neto2021, Tandra2021}, and \sctxt{svm}\cite{Ponte2017}.
     
     \item \textbf{Deep explanations}. Variations in \sctxt{dl} models allow the determination of explainable features by decomposing the decision into the contributions of the input features\cite{Montavon2017} or by inferring the transfer function between layers\cite{Bach2015}.
     
     \end{itemize}
     
\end{itemize}

To the best of our knowledge, this study represents the first attempt to explain industrial \sctxt{cf} estimations from the automatic classification of enterprise bank transactions. The proposed approach takes advantage of the different \sctxt{ml} models. Therefore, it follows a model-agnostic explainability strategy. Finally, an automatic validation of the explanation quality is provided.

\section{Methodology}
\label{architecture}

\begin{figure*}[!htpb]
 \centering
 \includegraphics[scale=0.37]{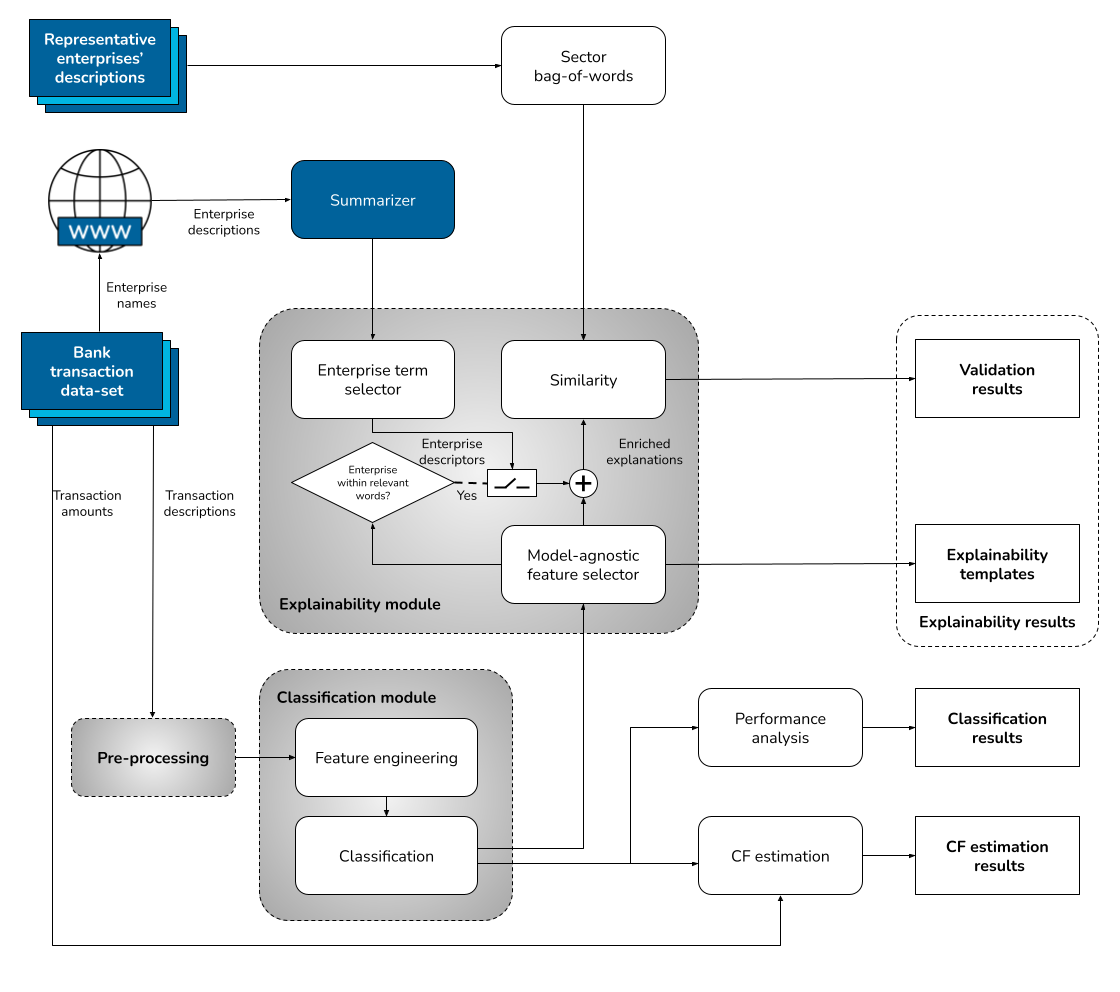}
 \caption{System architecture.}
 \label{fig:architecture}
\end{figure*}

Figure \ref{fig:architecture} illustrates the modular scheme of the proposed solution. The frames in white represent the elements within the processing pipeline, while the frames in blue represent external sources. Gray blocks correspond to higher-level tasks, as described in the independent subsections. This section aims to provide a conceptual perspective. Detailed implementations are described in Section \ref{evaluation}. 

In summary, the classification module labels the bank transactions used to estimate the \sctxt{cf}. Then, the explainability module automatically generates and validates the descriptions associated with the classifier decisions.

\subsection{Pre-processing}
\label{preprocessing}

The features used as input data for the classification task were engineered from textual bank transaction data. For this purpose, the text was processed using the following \sctxt{nlp} techniques:

\begin{itemize}

 \item \textbf{Numbers' removal}. Bank textual data usually contain quantitative information such as the receiver's bank account, receipt number, and product codes. These numbers are typically irrelevant for classification purposes because they are transaction specific.
 
 \item \textbf{Terms' reconstruction}. As bank descriptions are rather limited in length, relevant terms may be abbreviated or replaced with acronyms. Thus, these terms need to be expanded into natural language.
 
 \item \textbf{Removal of symbols and diacritic marks}. All symbols (\textit{e.g.}, asterisks, hyphens, \textit{etc.}), accents, diacritic marks, and diaereses were removed prior to text lemmatization.
 
 \item \textbf{Stop words and code removal}. Words with little semantic load, such as determiners, prepositions, general-usage verbs, and alphanumeric codes (\textit{e.g.}, customer identifiers), are removed.
 
 \item \textbf{Text lemmatization}. Finally, the remaining terms are split into tokens and converted into lemmas. 
 
\end{itemize}

\subsection{Classification module}
\label{classification_mod}

Once the processed bank transaction descriptions contain mostly semantically meaningful terms, the classification task is performed.

\subsubsection{Feature engineering}
\label{feat_eng}

Before training the \textsc{ml} models, the outcome of the pre-processing module is converted into vectors. Specifically, the resulting terms of each bank transaction description are transformed into wordgram elements, \textit{i.e.}, complete words, provided that our final goal is explainability. 

\subsubsection{Classification}
\label{classification}

Transactions are classified using learning models that fulfill two requirements: (\textit{i}) high classification performance of the target labels used for \sctxt{cf} estimation (see Section \ref{evaluation}), and (\textit{ii}) straightforward extraction of self-explainable features from the trained estimators to fill the explainability templates. Based on their suitability in explainability research in the literature, \sctxt{svm}, \sctxt{rf}, and Recursive Neural Networks ({\sctxt{rnn}}) were selected.

\subsection{Explainability module}
\label{explainability}

The goal of the explainability methodology is to provide a human operator with an in-depth understanding of the classification process and validate the corresponding relevant explanation terms with a metric of proximity between these terms and the descriptions of the \sctxt{cf} categories. In principle, the explanatory terms are those that are considered relevant during training by the model. 

However, due to the combined effect of pre-processing and feature engineering on the short bank transaction textual data, the explanations are enriched as follows:

\subsubsection{Enterprise term selection}
\label{bags}

Sometimes, the descriptions of transactions include explicit references to particular enterprises. By identifying these enterprises (see Section \ref{evaluation}), it is possible to retrieve their descriptions from the Internet, which are likely to be representative of their activity sector. These descriptions were pre-processed using the same method described in Section \ref{preprocessing}. The {\it summarizer} extracts all the nouns in the processed descriptions, and from these, the {\it enterprise term selector} takes the most representative ones, as detailed in Section \ref{explainability_dec}.

Additionally, the {\it similarity} calculation requires the collection of representative terms for each sector. Therefore, a bag-of-words is generated per sector using descriptions of the most representative enterprises (see Section \ref{evaluation}).

\subsubsection{Model-agnostic feature selection}
\label{feature_selection} 

Because the classification module employs different \sctxt{ml} models with particular internal structures, the system follows a model-agnostic approach. The latter method creates a local surrogate\cite{Burkart2021} model to select the explanatory terms for each bank transaction. The {\it model-agnostic feature selector} recursively analyzes the feature relevance by removing particular features (the deeper the impact, the higher the relevance). If any enterprise name is present in the initial batch of explanation terms for a bank transaction, those explanation terms are expanded using the descriptors of the enterprise, thanks to the {\it enterprise term selector}.

\subsubsection{Similarity}
\label{similarity}

Given the expanded explanation sets, the explainability module computes a similarity metric between the explanation set for each bank transaction and the bag-of-words of the economic sector, as selected by the classifier. Previous authors have also used contextual and semantic information to enhance explainability\cite{Kiefer2022,Rozanec2022}.

\begin{lstlisting}[caption={Original template in Spanish.}\label{lst:temp_spa}]
(*@\textit{La clasificación del movimiento} \textless transaction\_id\textgreater\ \textit{en la categoría} \textless output\_category\textgreater\ \textit{puede explicarse en order decreciente por los términos relevantes}: \textless term$_1$\textgreater\ ... \textless term$_n$\textgreater.@*)
\end{lstlisting}

\begin{lstlisting}[caption={Template translated to English.}\label{lst:temp_eng}]
(*@The classification of transaction \textless transaction\_id\textgreater\ into the category \textless output\_category\textgreater\ can be explained by relevant terms: (in decreasing order) \textless term$_1$\textgreater\ ... \textless term$_n$\textgreater.@*)
\end{lstlisting} 

\subsection{Carbon footprint estimation}
\label{carbon}

Once the transactions are classified, the proposed system automatically obtains their estimated \sctxt{cf} from the formulae of the sectors to which they are predicted to belong and the bank transaction amount, as described in Section \ref{carbon_dec}.

\section{Experimental Evaluation and Discussion}
\label{evaluation}

In this section, we present the experimental data-set and technical implementations.

\subsection{Experimental data-set}
\label{dataset}

The data-set is composed of 25,853 bank transactions issued by Spanish banks compiled by CoinScrap Finance SL\footnote{Available at \url{https://coinscrapfinance.com}, November 2022.}. Note that this data-set is comparable in size to that in our previous study on bank transaction classification\cite{Garcia-Mendez2020}.

It was downsampled using the \texttt{FuzzyWuzzy} Python library\footnote{Available at \url{https://pypi.org/project/fuzzywuzzy}, November 2022.} to keep only those entries sufficiently representative and distinguishable. Those samples with descriptions with a similarity greater than 90 \% were discarded. The downsampling process resulted in 2,619 transaction archetypes, with an average length of 10 words/73 characters.

The transactions are divided into three main categories: car and transport (\textit{automóvil y yransporte}), enterprise expenditures (\textit{gastos de empresa}), and commodities (\textit{suministros}), and several subcategories. Thus, a multi-class transformation\cite{Pereira2018,Tarekegn2021} process is applied to combine the main categories with their respective subcategories to map the following \sctxt{coicop} categories:

\begin{itemize}
 \item \textbf{Car and transport - gas stations (\textit{gasolineras})} (\sctxt{coicop} 7.2). Payments in gas stations.
 \item \textbf{Car and transport - private transport (\textit{transporte privado})} (\sctxt{coicop} 7.3). Payments in private transport services.
 \item \textbf{Car and transport - public transport (\textit{transporte público})} (\sctxt{coicop} 7.3). Purchase of public transportation tickets (buses and trains).
 \item \textbf{Car and transport - flights (\textit{vuelos})} (\sctxt{coicop} 7.3). Purchase of airline tickets.
 \item \textbf{Enterprise expenditures - parcel and courier (\textit{paquetería y mensajería})} (\sctxt{coicop} 8.1). Payment of public and private postal services.
 \item \textbf{Commodities - water bill (\textit{agua})} (\sctxt{coicop} 4.4). Water supply receipts.
 \item \textbf{Commodities - electricity bill (\textit{electricidad})} (\sctxt{coicop} 4.5). Receipt of energy supply.
 \item \textbf{Commodities - gas bill (\textit{gas})} (\sctxt{coicop} 4.5). Gas supply receipts.
\end{itemize}

Table \ref{tab:ori_data_set} shows the distribution of transactions by economic sector. Regarding description lengths, for instance, the category commodities - electricity bill has, on average, 16 words per description, while car and transport - private transport has only 6 words per description. 
Bank transaction pre-processing reduces the overall average description size to 7 words/50 characters.

\begin{table}[!htp]\centering
\caption{Distribution of samples in the data-set.}\label{tab:ori_data_set}
\begin{tabular}{ll}\toprule
\textbf{Category} & \textbf{Percentage} \\\midrule
Car and transport - gas stations & 23.18 \% \\
Car and transport - private transport & 10.84 \% \\
Car and transport - public transport & 9.00 \% \\
Car and transport - flights & 11.34 \% \\
Enterprise expenditures - parcel and courier & 7.25 \% \\
Commodities - water bill & 16.80 \% \\
Commodities - electricity bill & 16.15 \% \\
Commodities - gas bill & 5.38 \% \\
\bottomrule
\end{tabular}
\end{table}

\subsection{Implementations}
\label{discussion}

Experiments were performed on a computer with the following specifications:

\begin{itemize}
 \item \textbf{Operating System}. Ubuntu 20.04.3 LTS 64 bits
 \item \textbf{Processor}. Intel\@Xeon Platinum 8375C 2.9 GHz
 \item \textbf{RAM}. 64 \sctxt{gb} DDR4 
 \item \textbf{Disk}. 500 \sctxt{gb} SSD
\end{itemize}

For clarity, the corresponding architecture in Section \ref{architecture} is indicated for each implementation description.

\subsubsection{Pre-processing module (implementation of Section \ref{preprocessing})}
\label{preprocessing_dec}

Diacritic marks, numbers, identifiers, and codes were removed using regular expressions. The same technique is used to reconstruct common acronyms, such as \sctxt{s.l.} (\textit{Sociedad Limitada}, Limited Company) or \sctxt{e.s.} (\textit{estación de servicio}, gas station). Stop word removal (including general-usage verbs such as ``to be'' and ``do'') is based on the Spanish stop word list from the \texttt{NLTK} Python library\footnote{\label{ft:nltk}Available at \url{https://www.nltk.org}, November 2022.}. For tokenizing purposes, the same \texttt{NLTK} Python library\footref{ft:nltk} was used and the resulting tokens were lemmatized with the \texttt{spaCy} Python library\footnote{Available at \url{https://spacy.io}, November 2022.} using the {\tt es\_core\_news\_sm} model\footnote{Available at \url{https://spacy.io/models/es\#es_core_news_sm}, November 2022.}.

\subsubsection{Feature engineering module (implementation of Section \ref{feat_eng})}
\label{feat_eng_dec}

The selected classification models require different vectorization processes. For the \textsc{svc} and \textsc{rf} models, the \texttt{CountVectorizer}\footnote{Available at \url{https://scikit-learn.org/stable/modules/generated/sklearn.feature_extraction.text.CountVectorizer.html}, November 2022.} function from the \texttt{scikit-learn} Python library was used for wordgram extraction. After the preliminary tests, wordgrams (one word) and biwordgrams (two words) were extracted. The features were downsampled using \texttt{SelectPercentile}\footnote{\label{ft:sp}Available at \url{https://scikit-learn.org/stable/modules/generated/sklearn.feature_selectio.SelectPercentile.html}, November 2022.} function from the \texttt{scikit-learn} Python library to keep those with the highest correlation with the target variable. Prior knowledge led us to select the chi-squared score function\cite{DeArriba-Perez2020}.

For the \sctxt{lstm} model, the \texttt{Tokenizer}\footnote{Available at \url{https://www.tensorflow.org/api_docs/python/tf/keras/preprocessing/text/Tokenizer}, November 2022.} function from the \texttt{Keras} Python library was used. It converts text into sequences of token identifiers embedded in the network.

\subsubsection{Classification module (implementation of Section \ref{classification})}
\label{classification_dec}

The following models were used:

\begin{itemize}

 \item \textbf{Linear Support Vector Classification} (\sctxt{svc}). \texttt{Linear}-
 \texttt{SVC}\footnote{Available at \url{https://scikit-learn.org/stable/modules/generated/sklearn.svm.LinearSVC.html}, November 2022.} implementation from \texttt{scikit-learn} Python library.
 
 \item \textbf{Random Forest} (\sctxt{rf}). \texttt{RandomForestClassifier}\footnote{Available at \url{https://scikit-learn.org/stable/modules/generated/sklearn.ensemble.RandomForestClassifier.html}, November 2022.} implementation from the \texttt{scikit-learn} Python library.
 
 \item \textbf{Long Short-Term Memory} (\sctxt{lstm}). The \texttt{Sequential}\footnote{Available at \url{https://keras.io/api/models/sequential}, November 2022.} structure and \texttt{LSTMLayer}\footnote{\label{ft:lstm}Available at \url{https://keras.io/api/layers/recurrent_layers/lstm}, November 2022.} were implemented from the \texttt{Keras} Python library.
 
\end{itemize}

Hyperparameter selection for the \sctxt{svc} and \sctxt{rf} models was performed using the \texttt{GridSearchCV}\footnote{Available at \url{https://scikit-learn.org/stable/modules/generated/sklearn.model_selection.GridSearchCV.html}, November 2022.} function of the \texttt{scikit-learn} Python library. Listings \ref{hyperparam_svc} and \ref{hyperparam_rf} detail the hyperparameter ranges and the final choices, respectively.

\begin{lstlisting}[frame=single,caption={Hyperparameter selection for \sctxt{svc} (best values in bold).}, label={hyperparam_svc},emphstyle=\textbf,escapechar=ä]
class_weight = [None, ä\textbf{balanced}ä],
loss = [ä\textbf{hinge}ä, squared_hinge],
max_iter = [50, ä\textbf{100}ä, 250, 500, 1000],
multi_class = [ovr, ä\textbf{crammer\_singer}ä],
tol = [ä\textbf{\expnumber{1}{-10}}ä, ä\expnumber{1}{-9}ä, ä\expnumber{1}{-8}ä, ä\expnumber{1}{-7}ä, ä\expnumber{1}{-6}ä, ä\expnumber{1}{-5}ä, ä\expnumber{1}{-4}ä, ä\expnumber{1}{-3}ä,
ä\expnumber{1}{-2}ä],
penalty = [ä\textbf{l2}ä],
C = [ä\expnumber{1}{-4}ä, ä\expnumber{5}{-3}ä, ä\expnumber{1}{-3}ä, ä\expnumber{5}{-2}ä, ä\expnumber{1}{-2}ä, ä\expnumber{5}{-1}ä, ä\textbf{\expnumber{1}{-1}}ä, 1]]
\end{lstlisting}

\begin{lstlisting}[frame=single,caption={Hyperparameter selection for \sctxt{rf} (best values in bold).}, label={hyperparam_rf},emphstyle=\textbf,escapechar=ä]
n_estimators = [50, 100, 250, ä\textbf{500}ä, 1000, 
2000],
max_depth = [10, 25, 50, ä\textbf{100}ä, 200],
max_leaf_nodes = [50, 100, ä\textbf{250}ä, 500],
criterion = [ä\textbf{gini}ä, entropy]
\end{lstlisting}

The configuration used for the \sctxt{lstm} model included the \texttt{SpatialDropout}\footnote{Available at \url{https://keras.io/api/layers/regularization_layers/spatial_dropout1d}, November 2022.} layer (equivalent to \texttt{Select}-
\texttt{Percentile}\footref{ft:sp}). The final dropout percentage applied prior to \texttt{LSTMLayer}\footref{ft:lstm} was 20 \% of the tensors. 
The drop percentage of \texttt{LSTMLayer}\footref{ft:lstm} was also 20 \%.

\subsubsection{Explainability module (implementation of Section \ref{explainability})}
\label{explainability_dec}

The explainability methodology comprises two complementary processes: (\textit{i}) the generation of explanations based on explainability templates and (\textit{ii}) the validation of these explanations in terms of their consistency compared with human knowledge about the target sectors.

The similarity metric of the validation process uses the bag-of-words from the target sectors and descriptors of the enterprises in the experimental data-set as input. For the former element, CoinScrap Finance \sctxt{s.l.} provided a corporate lexicon\footnote{Available at \url{https://docs.google.com/spreadsheets/d/1Tq2l9An6DybVTHig_5O_5_KN-0VucgjSFT8wGQluahc/edit?usp=sharing}, November 2022.} created from the descriptions of the top six enterprises of each target sector, with ten representative nouns and five representative verbs each. Conversely, Spanish companies' names\footnote{Available at \url{https://guiaempresas.universia.es/localidad/MADRID}, November 2022} and their descriptions\footnote{Available at \url{https://docs.google.com/spreadsheets/d/1SNT4avp9ki4beD6tYCH27zE6FQpsQUXTdH5vuLVF0yc/edit?usp=sharing}, November 2022.} were extracted from the Internet. 

The \textit{enterprise term selector} chooses up to ten terms from each summarized description of an enterprise in the data-set. The \textit{summarizer} followed the same steps as the pre-processing module by removing stop words, common verbs, numbers, and codes. The system creates a list of words for each enterprise from the resulting list of lemmas. If the list contains more than ten terms, only the ten most frequent terms are retained. 

The \textit{model-agnostic feature selector} performs recursive feature selection tests using the \texttt{\textsc{lime}}\footnote{Available at \url{https://github.com/marcotcr/lime}, November 2022.} Python library given its wide acceptance in the literature\cite{Burkart2021}. As previously explained, the previous features are enriched with the enterprise descriptors obtained by the \textit{enterprise term selector} in case the bank description contains the name of a company to generate the explanation sets.

For similarity metrics between groups of terms, we considered two different approaches: \textit{(i)} Jaccard similarity as a baseline\cite{Jain2017, Singh2021}, and \textit{(ii)} our own sophisticated metric based on lexical and semantic proximity\cite{DeArriba-Perez2022}. The cosine distance was discarded provided that the terms in the descriptions had no logical ordering. 
Using a similarity metric, the system calculates the similarities between enriched bank transaction explanations and the bag-of-words of the target sectors so that the highest similarity can be expected between an enriched explanation of a bank transaction and its target sector according to the classification module.

\subsubsection{Carbon footprint module (implementation of Section \ref{carbon_dec})}
\label{carbon_dec}

The \sctxt{cf}, that is, the \sctxt{ghg} emissions associated with a transaction, is directly related to the transaction amount. The conversion estimate depends on the sector:

\begin{itemize}

 \item \textbf{Car and transport - gas stations}. \sctxt{co\textsubscript{2}} emissions $CF_{gs}$ depend on fuel volume and the emission factor of the fuel $\epsilon_f$. As bank transactions do not include the type of fuel, the emission factor averages the emissions of gasoline and diesel. The volume is derived from the payment amount $p$ and the average price per liter $avp_f$ of fuel at transaction time.
 \begin{equation}
 CF_{gs}=\frac{p}{avp_f}\cdot\epsilon_f
 \end{equation}
 
 \item \textbf{Car and transport - private/public transport}. For private transport, it is necessary to first distinguish between taxi payments and other private services. We applied these keywords for this purpose. Each of these alternatives has its own emission factor, $\epsilon_t$ and $\epsilon_c$, respectively. Distances are calculated from the average prices per kilometer of the region $avp_t$ (for taxis) and $avp_c$ (for private companies) and the amount paid $p$. Prices are obtained from official and private pricing lists. 
 
 For taxis:
 
 \begin{equation}
 CF_{taxi}=\frac{p}{avp_t}\cdot\epsilon_t
 \end{equation}
 
 For private companies, $CF_{comp}$ is defined similarly as $avp_c$ and $\epsilon_c$. For public transport, the Spanish Transport Ministry publishes average references for the price per kilometer\footnote{Available at \url{https://www.mitma.gob.es/transporte-terrestre/observatorios/observatorios-y-estudios} (Spanish), November 2022.} $avp_{pt}$. \sctxt{co\textsubscript{2}} emissions depend on the travel distance. The emission factor for the transportation means considered is $\epsilon_{pt}$.
 \begin{equation}
 CF_{pt}=\frac{p}{avp_{pt}}\cdot\epsilon_{pt}
 \end{equation}
 
 \item \textbf{Car and transport - flights}. In this sector, the price per kilometer $avp_{fl}$ must be averaged, as it varies depending on the airline and plane model. Given this estimate and the payment amount $p$, we calculate the \sctxt{co\textsubscript{2}} emission from the corresponding travel distance and the emission factor for a commercial aircraft $\epsilon_{fl}$.
 \begin{equation}
 CF_{fl}=\frac{p}{avp_{fl}}\cdot\epsilon_{fl}
 \end{equation}
 
 \item \textbf{Enterprise expenditures - parcel and courier}. Both private companies and public entities record parcel transport costs per kilometer $avp_{pc}$ on a yearly basis. From these data and the amount $p$, it is possible to estimate the shipment distance and, therefore, its \sctxt{cf} from the emission factor $\epsilon_{pc}$.
 \begin{equation}
 CF_{pc}=\frac{p}{avp_{pc}}\cdot\epsilon_{pc}
 \end{equation}
 
 \item \textbf{Commodities - water bill}. Unlike the rest of the bank transactions, for water bills, we do not calculate \sctxt{ghg} emissions but the total consumption of water $TWC$, which depends on the average price of the service $avp_w$ and the amount paid $p$.
 \begin{equation}
 TWC=\frac{p}{avp_{w}}
 \end{equation}
 
 \item \textbf{Commodities - electricity/gas bill}. The daily price per kWh $kwp_i$ for the $i$-day of the last month is publicly available. The consumption of electricity from a receipt with amount $p$ is estimated from the average price in the previous month. From the emission factor $\epsilon_{e}$ for electricity:
 \begin{equation}
 CF_{e}=\frac{p}{avp_{e}}\cdot\epsilon_{e}=\frac{p}{\frac{1}{m}\sum_{i=1}^{m}kwp_i}\cdot\epsilon_{e}
 \end{equation}
 where $m$ denotes the number of days in the previous month. $CF_{g}$ is defined similarly from $\epsilon_{g}$. 
 
\end{itemize}

From the predicted class of transactions and their amounts, the system presents the users with the estimated volume of \sctxt{ghg} associated with each transaction. Table \ref{tab:carbon_results} presents examples of transactions and their corresponding \sctxt{co\textsubscript{2}} emissions in kilograms. Table \ref{tab:water_results} presents an example of water consumption estimated from a water bill transaction.

\begin{table*}[!htp]
 \centering
\caption{Samples of \textsc{cf} estimation results.}\label{tab:carbon_results}
 \begin{tabular}{lcccc}
 \toprule
\textbf{Description} & \textbf{Amount (\EUR)} & \textbf{Predicted sector} & \textbf{Parameters} & \textbf{\sctxt{co\textsubscript{2}} emission (kg)} \\ \midrule
\multirow{2}{*}{BALLENOIL ALBAL} & \multirow{2}{*}{80.0} & Car and transport & avp\textsubscript{f}=\SI[per-mode=symbol]{1.83}{\EUR\per\liter} & \multirow{2}{*}{102.733} \\
&&Gas stations&\textepsilon\textsubscript{f}=\SI[per-mode=symbol]{2.35}{\kgc\per\liter}&\\
\midrule
\multirow{2}{*}{LIC {[}NUM{]} TAXI MADRID} & \multirow{2}{*}{10.1} & Car and transport & avp\textsubscript{t}=\SI[per-mode=symbol]{2.02}{\EUR\per\km} & \multirow{2}{*}{0.855} \\
&&Private transport&\textepsilon\textsubscript{t}=\SI[per-mode=symbol]{0.17}{\kgc\per\km}&\\
\midrule
\multirow{2}{*}{Tj-renfe virtual internet} & \multirow{2}{*}{142.6} & Car and transport & avp\textsubscript{pt}=\SI[per-mode=symbol]{0.15}{\EUR\per\km} & \multirow{2}{*}{33.273} \\
&&Public transport&\textepsilon\textsubscript{pt}=\SI[per-mode=symbol]{0.035}{\kgc\per\km}&\\
\midrule
\multirow{2}{*}{COMPRA TARJ. {[}NUM{]} Ryanair-Madrid} & \multirow{2}{*}{34.68} & Car and transport & avp\textsubscript{fl}=\SI[per-mode=symbol]{0.05}{\EUR\per\km} & \multirow{2}{*}{133.171} \\
&&Flights&\textepsilon\textsubscript{fl}=\SI[per-mode=symbol]{0.192}{\kgc\per\km}&\\
\midrule
\multirow{2}{*}{SE CORREOS Y TELEGRAFOS S (VILLENA)} & \multirow{2}{*}{29.0} & Enterprise expenditures & avp\textsubscript{pc}=\SI[per-mode=symbol]{1.3}{\EUR\per\km} & \multirow{2}{*}{3.525} \\
&&Parcel and courier&\textepsilon\textsubscript{pc}=\SI[per-mode=symbol]{0.158}{\kgc\per\km}&\\
\midrule
\multirow{2}{*}{FACTURA DE GAS PM {[}NUM{]} {[}NUM{]}} & \multirow{2}{*}{48.04} & Commodities & avp\textsubscript{g}=\SI[per-mode=symbol]{0.1398}{\EUR\per\kwh} & \multirow{2}{*}{69.757} \\
&&Gas bill&\textepsilon\textsubscript{g}=\SI[per-mode=symbol]{0.203}{\kgc\per\kwh}&\\
\midrule
\multirow{2}{*}{RECIBO IBERDROLA CLIENTES, S.A.U RECIBO {[}NUM{]}} & \multirow{2}{*}{23.0} & Commodities & avp\textsubscript{e}=\SI[per-mode=symbol]{0.098}{\EUR\per\kwh} & \multirow{2}{*}{58.673} \\
&&Electricity bill&\textepsilon\textsubscript{e}=\SI[per-mode=symbol]{0.25}{\kgc\per\kwh}&\\
\bottomrule
\end{tabular}
\end{table*}

\begin{table*}[!htp]
 \centering
\caption{Sample of water consumption results.}\label{tab:water_results}
 \begin{tabular}{lcccc}
 \toprule
\textbf{Description} & \textbf{Amount (\EUR)} & \textbf{Predicted sector} & \textbf{Parameters} & \textbf{Water consumption (L)} \\\midrule
RECIBO AGUA-[NUM]-BO. & 50.11 & Commodities - water bill &avp\textsubscript{w}=\SI[per-mode=symbol]{0.0017}{\EUR\per\liter} & 29304.094 \\
\bottomrule
\end{tabular}
\end{table*}

\subsection{Classification results}

$K$-fold cross-validation is a common strategy for proper validation of prediction results\cite{Jiang2020}. In particular, we applied a 10-fold cross-validation,
as implemented with the \texttt{StratifiedKFold}\footnote{Available at \url{https://scikit-learn.org/stable/modules/generated/sklearn.feature_selection.SelectPercentile.html}, November 2022.} function from the \texttt{scikit-learn} Python library, to calculate average accuracy, precision, recall, and training times. Table \ref{tab:class_res} presents the results for the \sctxt{svc}, \sctxt{rf}, and \sctxt{lstm} models.

\begin{table}[!htp]\centering
\caption{Classification results.}\label{tab:class_res}
\begin{tabular}{ccccc}\toprule
\multirow{2}{*}{\textbf{Model}} & \multirow{2}{*}{\textbf{Accuracy}} & \multirow{2}{*}{\textbf{Precision}} & \multirow{2}{*}{\textbf{Recall}} & \textbf{Training}\\
&&&&\textbf{time (s)}\\
\midrule
\sctxt{svc} & 93.72 \% & 94.36 \% & 92.34 \% & 0.532 \\
\sctxt{rf} & 89.18 \% & 90.24 \% & 86.36 \% & 20.33 \\
\sctxt{lstm} & 92.34 \% & 92.37 \% & 90.97 \% & 399.09\\
\bottomrule
\end{tabular}
\end{table}

\sctxt{svc} and \sctxt{lstm} achieved over 90 \% of accuracy. \sctxt{svc} is the most time-efficient model. Regarding the training time, \sctxt{lstm} was the most time-consuming. \sctxt{rf} is an intermediate alternative with slightly lower performance. Consequently, the best model, considering the performance-time tradeoff, was \sctxt{svc}, but the classification performance of the three models selected was similar.

\subsection{Explainability performance results}

The \sctxt{rf} model was used as the baseline because of its inferior performance, whereas \sctxt{svc} was selected as the target classifier.

Table \ref{tab:sim_comp} shows the percentage of explanations for the \sctxt{rf} baseline model that could be validated directly. An explanation is considered to be directly ``validated'' when the sector closest to a bank transaction explanation is predicted by the classifier.

\begin{table}[!htp]\centering
\caption{Directly validated explanations for the \sctxt{rf} classifier.}\label{tab:sim_comp}

\begin{tabular}{lc}\toprule
\label{tab:validated}
\textbf{Metric} & \textbf{Validated} \\ \midrule
Jaccard & 34.85 \% \\
Linguistic proximity metric\cite{DeArriba-Perez2022} & 46.89 \% \\
\bottomrule
\end{tabular}
\end{table}

As shown in Table \ref{tab:validated}, Jaccard similarity results in a lower percentage of directly validated explanations. Therefore, in the rest of the experiments, our linguistic metric\cite{DeArriba-Perez2022} was used. This metric is well-suited to our goal, as it requires fewer terms per explanation than the Jaccard distance to detect similarity, and unlike the cosine distance, it does not rely on term ordering. The differences between the lists of terms in the bank transaction explanation sets generated for both models were analyzed. For \sctxt{rf}, each explanation set contained 7.67 words on average, while 8.19 words on average in the case of \sctxt{svc}. Similarities between pairs of explanation sets were then computed, resulting in an overall average similarity of 0.79. Note that the similar performances of both classification methods seem to be related to the similarity of their explanation sets. Thus, the explanation potential is consistent with the classification performance.

Table \ref{tab:sim_res} shows the explanatory performance of both the models. For explanations that were neither obvious nor directly validated, we performed a second in-depth analysis to check their trustworthiness in a human operator. We divided them by manual inspection into ``coherent'' (when the human operator considered that the explanation was correct given the predicted sector) and ``ambiguous'' (when the human operator could not determine from the explanation itself the sector that was predicted by the classifier). Those ``coherent'' explanations that contained the name of a company of the target sector are obviously satisfactory and, as such, they were marked as ``obvious''. Finally, we considered ``empty'' those explanations whose similarity to all sectors is zero. This may be due to the fact that \texttt{\textsc{lime}} fails to detect any representative term or no selected explanation term is representative enough (\textit{i.e.}, simple alphanumerical codes). Therefore, the lower the percentage of ambiguous and empty explanations, the higher the trustworthiness.

\begin{table}[!htp]\centering
\caption{Explanation performance.}\label{tab:sim_res}
\begin{tabular}{cccc|cc}\toprule

\textbf{Model} & \multicolumn{3}{c}{\bf Satisfactory} & \multicolumn{2}{c}{\bf Unsatisfactory} \\

\midrule

 & \textbf{Validated} & \textbf{Obvious} & \textbf{Coherent} & \textbf{Empty} & \textbf{Ambiguous} \\

\midrule
\sctxt{svc} & 48.13 \% & 9.96 \% & 15.77 \% & 12.79 \% & 13.35 \%\\
\sctxt{rf} & 46.89 \% & 9.54 \% & 15.35 \% & 13.28 \% & 14.94 \%\\
\bottomrule
\end{tabular}
\end{table}

The explanation performances were similar, which resulted from the use of a model-agnostic feature selector and the similar classification performance of both models. Satisfactory explanations exceeded 70 \%, of which approximately 60 \% could be automatically ``validated". Regarding unsatisfactory explanations (``empty" and ``ambiguous"), only over 12 \% are ``empty" and offer no information to a human operator.

\begin{figure}
 \centering
 \begin{subfigure}[b]{0.45\textwidth}
 \centering
 \includegraphics[width=\textwidth]{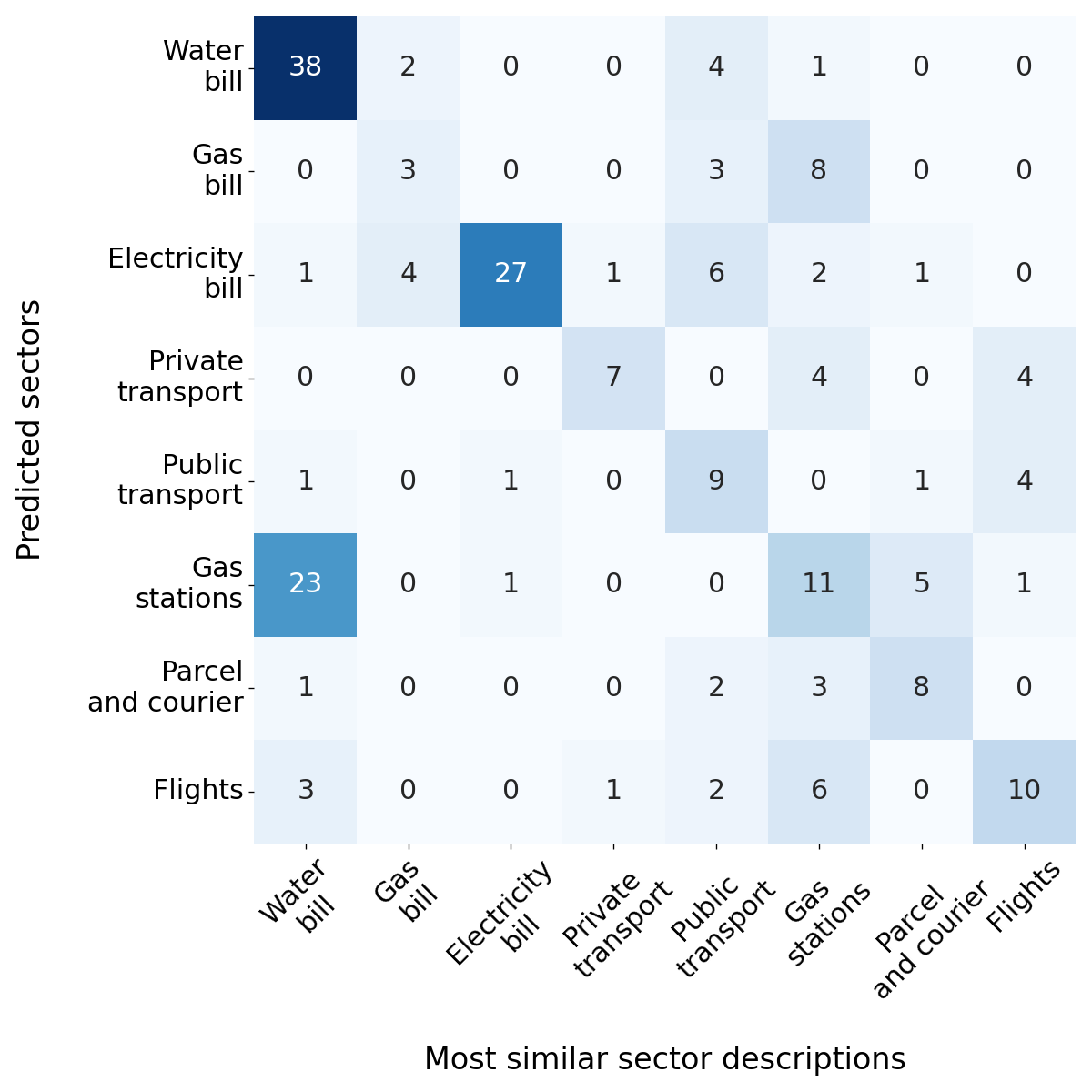}
 \caption{RF similarity confusion matrix.}
 \label{fig:RF_conf}
 \end{subfigure}
 \hfill
 \begin{subfigure}[b]{0.45\textwidth}
 \centering
 \includegraphics[width=\textwidth]{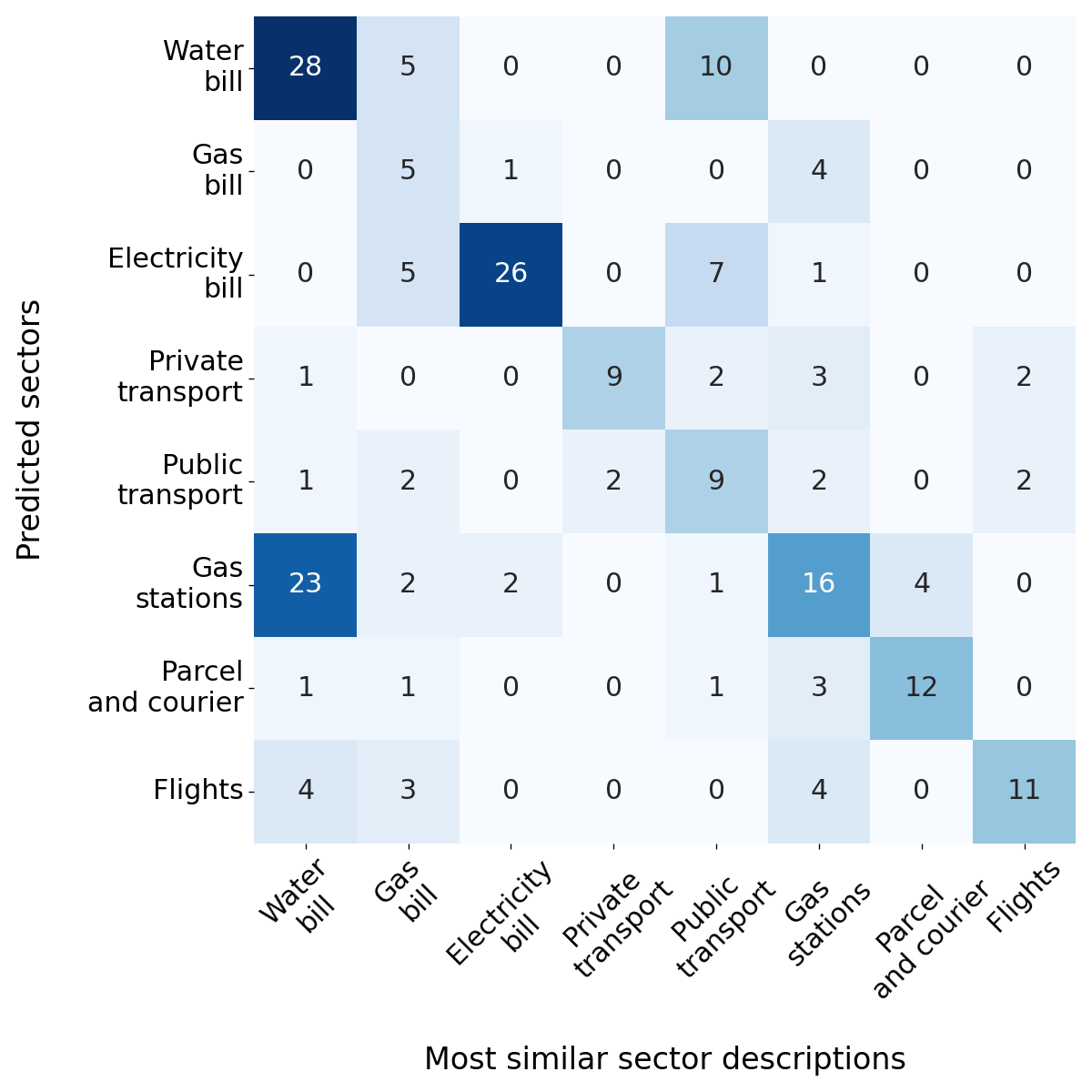}
 \caption{SVC similarity confusion matrix.}
 \label{fig:SVC_conf}
 \end{subfigure}
 
 \caption{Confusion matrices, predicted sectors versus most similar sector descriptions.}
 \label{fig:conf_matr}
\end{figure}

Figure \ref{fig:conf_matr} illustrates the confusion matrices of the predicted sectors versus the most similar sector descriptions for the direct validation. The main deviation occurred for gas stations, the category with the shortest explanations, with only four words on average. These are frequently closest to the water bill sector description. Other common deviations exist between gas stations and the gas bill, and between the three categories of commodities.

Most of these deviations do not correspond to unsatisfactory results from the perspective of a human operator. For example, let us consider an explanation of a bank transaction that was predicted to belong to the commodities - electricity bill and was closer to car and transport - public transport. The explanation set contained the relevant terms `energia', `referencia', `recibo referencia', `recibo', `energy', `nxs', `nexus', `mandato nxs', `nexus energia' and `referencia mandato'. It includes several instances of the meaningful terms \textit{energía} and \textit{energy} that, in the Spanish context, are directly related to the electricity sector from the perspective of the human operator.

The system finally presents explanations by following the template in Listing \ref{lst:temp_eng}. Some examples are:

\begin{itemize}
\item \textbf{Car and transport - gas stations.} \textit{The classification of transaction 423 into the category car and transport - gas stations can be explained by relevant terms (in decreasing order)}: cedipsa, service (\textit{servicio}), station (\textit{estacion}), gas station (\textit{estacion servicio}), payment (\textit{pago}), cedipsa payment (\textit{pago} cedipsa).

\item \textbf{Car and transport - public transport.} \textit{The classification of transaction 895 into category car and transport - public transport can be explained by relevant terms (in decreasing order)}: renfe, madrid, travelers (\textit{viajeros}), renfe card (\textit{tarjeta} renfe), renfe travelers (\textit{viajeros} renfe), purchase (\textit{compra}), travelers app (\textit{viajeros} app), app, dev, card (\textit{tarjeta}).

\item \textbf{Enterprise expenditures - parcel and courier.} \textit{The classification of transaction 1269 into category enterprise expenditures - parcel and courier can be explained by relevant terms (in decreasing order)}: mail (\textit{correos}), mail payment (\textit{pago correos}), purchase (\textit{compra}), purchase payment (\textit{pago de compra}), mail leganes (\textit{correos leganes}), card (\textit{tarjeta}).

\item \textbf{Commodities - water bill.} \textit{The classification of transaction 1514 into category commodities - water bill can be explained by relevant terms (in decreasing order)}: water (\textit{agua}), water receipt (\textit{recibo agua}), receipt (\textit{recibo}), reference order (\textit{referencia mandato}), reference (\textit{referencia}), order (\textit{mandato}), receipt reference (\textit{referencia recibo}).

\end{itemize}

In these examples, the lists of relevant terms in the explanations are highly related to the respective sectors. These include \textit{electricidad} (electricity), \textit{gas} (gas), \textit{agua} (water), and \textit{estacion servicio} (gas station). Two of the explanations are obvious because they contain the names of companies offering the services (Cedipsa and Renfe), but they also contain other highly informative words. There are other generalist terms, such as \textit{recibo} (receipt), \textit{compra} (purchase), and \textit{tarjeta} (card). Although they do not occupy the first positions in their respective lists, they are less relevant than sector-specific terms.

\subsection{Comparison with prior work}
\label{comparison}

The \sctxt{cf} estimation has recently attracted significant commercial interest. However, there are few automatic solutions based on bank transaction classification in the literature owing to its novelty.

Although a few previous studies have applied bank transaction classification to industrial use cases, the classification performance achieved by other researchers on different finance-related problems is illustrative.

E. Folkestad \textit{et al.} (2017)\cite{Folkestad2017} exploited data from DBpedia\footnote{Available at \url{https://es.dbpedia.org}, November 2022.} and Wikidata\footnote{Available at \url{https://www.wikidata.org}, November 2022.} for bank transaction classification. They reported 83.48 \% accuracy using Logistic Regression (\textsc{lr}) (10.24 \% less than with our approach). Moreover, E. Vollset \textit{et al.} (2017)\cite{Vollset2017} augmented corporate data with external semantic resources to improve bank transaction classification. They obtained 92.97 \% accuracy also with \textsc{lr} (0.75 \% less than with our approach).

The \sctxt{nlp}-based budget management solution by S. Allegue \textit{et al.} (2021)\cite{Allegue2021} obtained similar results to our approach with an Adaptive Random Forest model, with a difference of only 1 \% in precision.

The non \sctxt{nlp}-based \sctxt{svm} solution for cash flow prediction for small \& medium enterprises by D. Kotios \textit{et al.} (2022)\cite{Kotios2022} attained a precision that was only 0.2 \% higher than ours, after trying many other algorithms.

Given the similar performance of the existing solutions, some contributions directly focus on the problem description. This is the case of the {\it Svalna} app by D. Andersson (2020)\cite{Andersson2020}, an automatic carbon footprint estimation application based on users' transactions and environmental data from governmental agencies.

This apparent intrinsic high separability of the problem is consistent with our own results with the three different classification methodologies. Because the focus of our contribution is on classification explainability and given the small gap between methodologies in this and other works, and despite the advantages of \sctxt{rf} for self-explainability\cite{Wanner2021}, we have applied a model-agnostic explanation methodology.

\section{Conclusions}
\label{conclusions}

In this study, a novel explainable solution for automatic industrial \sctxt{cf} estimation from bank transactions is proposed, addressing the lack of transparent decision explanation methodologies for this problem. The explanation is especially important to trust the outcome of automatic processes, for them to replace more expensive alternatives, such as consultancy analytics. Indeed, even though automatic explainability has not been tackled in this domain, the study of the state of the art has also revealed that there are no previous works or existing commercial solutions for automatic industrial \sctxt{cf} estimation based on bank transactions.

The original data source includes more than 25,000 bank transactions. It was annotated for classification using \sctxt{coicop} categories. 

The classification methodology for bank transactions followed a supervised learning strategy by combining \sctxt{ml} with \sctxt{nlp} techniques based on our approach in\cite{Garcia-Mendez2020}. The widely used \sctxt{svm}, \sctxt{rf}, and \sctxt{lstm} models achieve satisfactory performance levels of 90 \% for all metrics, which is consistent with the results reported by other authors in the literature.

The agnostic explanation methodology extracts a set of relevant words for the classifier, and this explanation set is then validated with a similarity metric by comparing the set with the descriptions of carbon-intensive activity sectors. Despite the scarcity of content in industrial bank transaction descriptions, over 70 \% of the explanations are satisfactory to a human operator, and 60 \% have been automatically validated from company descriptions of the target sectors. Only 15 \% of the explanations were ambiguous, and there is a margin for improvement in the rest (which we tag as ``empty'') if side information on alphanumeric codes of industrial activity is provided. We consider these results encouraging for further study on the automatic explainability of \sctxt{cf} estimation in industrial sectors.

In summary, the highlights of this study are as follows:

\begin{itemize}

\item The main contribution of this study is a novel solution for automatic industrial \sctxt{cf} estimation from bank transactions based on supervised \textsc{ml} and \textsc{nlp} techniques.

\item The performance of the underlying bank transaction classification methodology is comparable to that of other researchers\cite{Allegue2021,Kotios2022}.

\item An experimental data-set composed of more than 25,000 bank transactions was used.

\item Over 70 \% of the natural language explanations automatically generated with a model-agnostic approach are satisfactory for end users. Of these, 60 \% have been automatically validated. Less than 15 \% are ambiguous.

\end{itemize}

Regarding the limitations of this study, the supervised classification methodology requires manual annotation of bank transactions for training purposes. In addition, the categories for \sctxt{cf} estimation could change, depending on the activity sector. We chose a well-established reference, but finer details may be required to account for business-specific expenses.

In future work, we plan to extend this research to other main languages, enrich explanations with complementary enterprise information, and study the effect of hierarchical methodologies on categorization by leveraging the relations between target classes. We also plan to move towards a semi-supervised approach by combining the current solution with a rule scheme, such as those proposed by other authors\cite{Kotios2022}. Another possible line of research is the comparison of the model-agnostic approach to explainability with model-specific methodologies.

\bibliography{mendeley.bib}{}
\bibliographystyle{IEEEtran}

\begin{IEEEbiography}[{\includegraphics[width=1in,height=1.25in,clip,keepaspectratio]{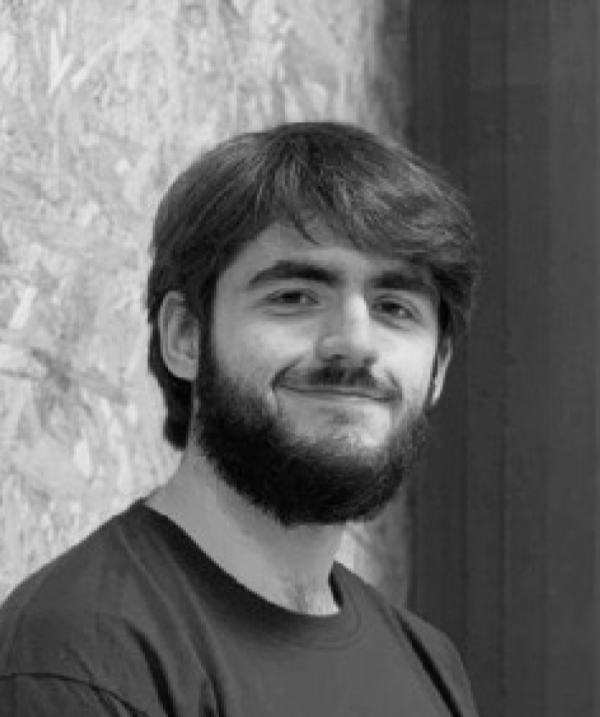}}]{Jaime González-González} received the B.S. degree in telecommunication technologies engineering in 2020 and the M.S. degree in telecommunication engineering in 2022, both from University of Vigo, Spain. He is currently a researcher and Ph.D. candidate in the Information Technologies Group at the University of Vigo. His research interests include the development of Machine Learning solutions for automatic text classification and Augmentative and Alternative Communication.
\end{IEEEbiography}

\begin{IEEEbiography}[{\includegraphics[width=1in,height=1.25in,clip,keepaspectratio]{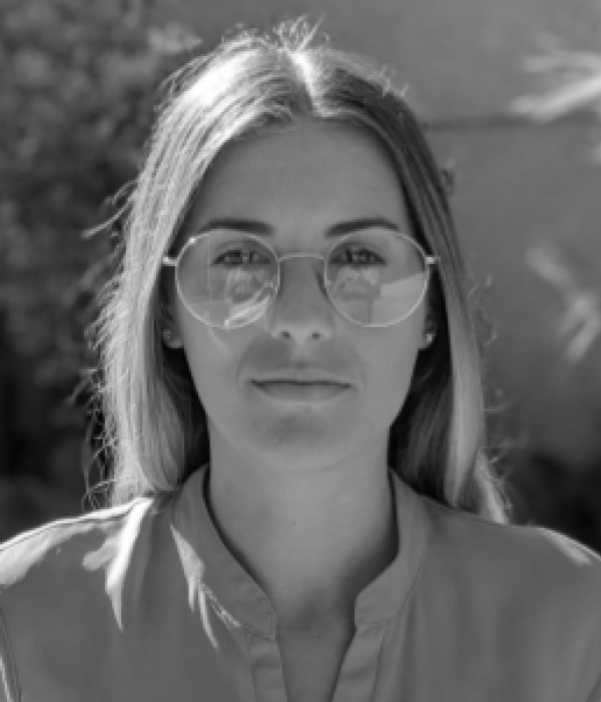}}]{Silvia García-Méndez} received the Ph.D. degree in Information and Communication Technologies from University of Vigo in 2021. Since 2015, she has been working as a researcher with the Information Technologies Group at the University of Vigo. She is currently collaborating with foreign research centers as part of her postdoctoral stage. Her research interests include Natural Language Processing techniques and Machine Learning algorithms.
\end{IEEEbiography}

\begin{IEEEbiography}[{\includegraphics[width=1in,height=1.25in,clip,keepaspectratio]{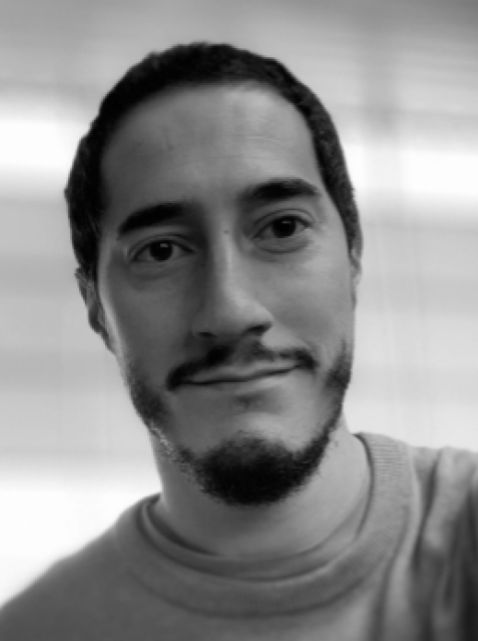}}]{Francisco de Arriba-Pérez} received the B.S. degree in telecommunication technologies engineering in 2013, the M.S. degree in telecommunication engineering in 2014, and the Ph.D. degree in 2019 from University of Vigo, Spain. He is currently a researcher in the Information Technologies Group at the University of Vigo, Spain. His research includes the development of Machine Learning solutions for different domains like finance and health.
\end{IEEEbiography}

\begin{IEEEbiography}[{\includegraphics[width=1in,height=1.25in,clip,keepaspectratio]{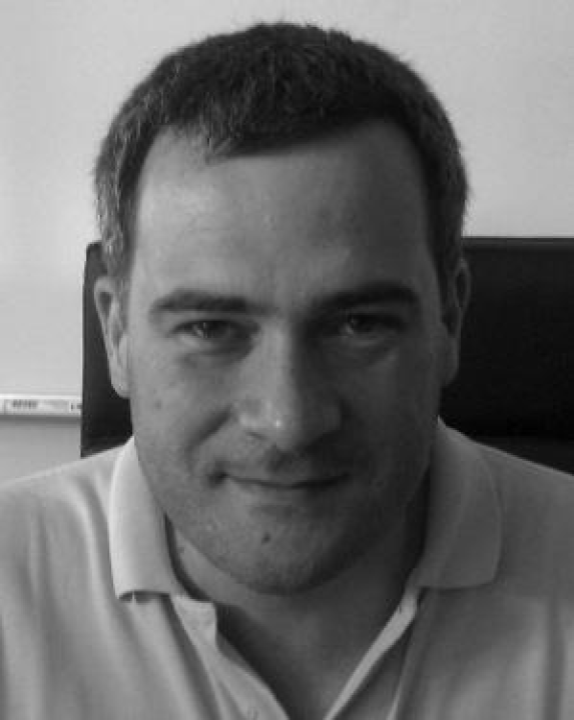}}]{Francisco J. González-Castaño} received the B.S. degree from University of Santiago de Compostela, Spain, in 1990, and the Ph.D. degree from University of Vigo, Spain, in 1998. He is currently a full professor at the University of Vigo, Spain, where he leads the Information Technologies Group. He has authored over 100 papers in international journals in the ﬁelds of telecommunications and computer science and has participated in several relevant national and international projects. He holds three U.S. patents.
\end{IEEEbiography}

\begin{IEEEbiography}[{\includegraphics[width=1in,height=1.25in,clip,keepaspectratio]{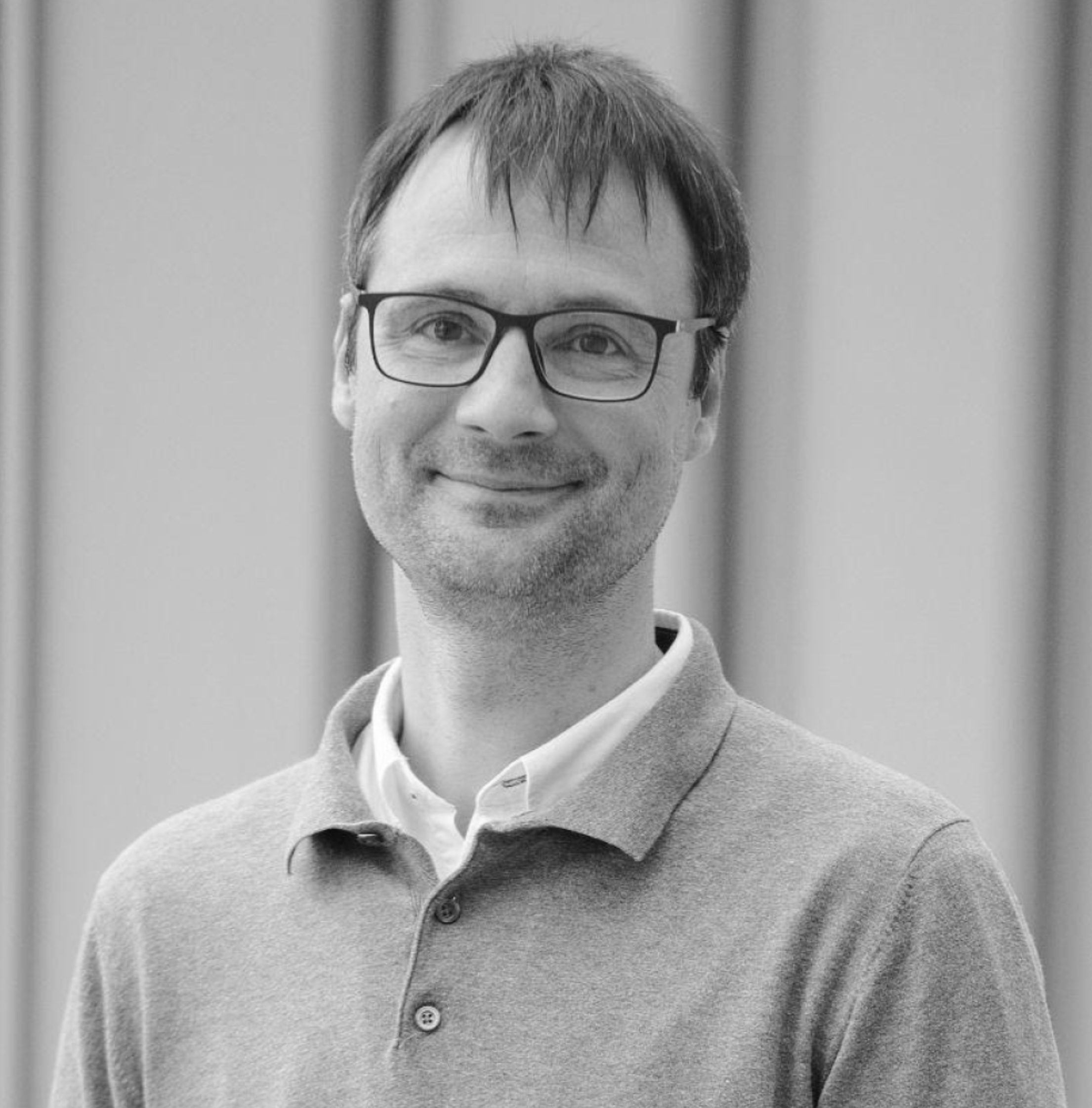}}]{Óscar Barba-Seara} received a B.S. degree in computer science and an M.S. degree in superior computer engineering from the University of Vigo, and an M.S. degree in e-commerce from the Pontifical University of Salamanca. He is currently pursuing a Ph.D. at the University of Vigo. He has worked as a CTO or Technical Manager in several IT projects with more than 14 years of experience in the public and private sectors and in IT integrations with international corporations such as Mapfre, Caser, Abanca, Evo Banco, Vodafone, and AON. He is currently a CTO of two initiatives in the fintech sector involving Machine Learning research for the analysis of financial and market-related texts.

\end{IEEEbiography}

\EOD

\end{document}